\documentclass{article}

\usepackage[nonatbib, final]{nips_2016}

\usepackage[utf8]{inputenc} 
\usepackage[T1]{fontenc}    
\usepackage{url}            
\usepackage{booktabs}       
\usepackage{amsfonts}       
\usepackage{nicefrac}       
\usepackage{microtype}      
\usepackage{subfigure}
\usepackage{graphicx}
\usepackage{comment}
\usepackage{soul}
\usepackage{multirow}
\usepackage{array}
\usepackage{amsmath}
\usepackage{bbm}
\usepackage[export]{adjustbox}
\usepackage[perpage]{footmisc}

\newcolumntype{P}[1]{>{\centering\arraybackslash}p{#1}}
\newcolumntype{M}[1]{>{\centering\arraybackslash}m{#1}}
\newcolumntype{N}{@{}m{0pt}@{}}

\title{Known Unknowns:\\Uncertainty Quality in Bayesian Neural Networks}

\newcommand*\samethanks[1][\value{footnote}]{\footnotemark[#1]}

\author{
   Ramon Oliveira\thanks{The first two authors contributed equally to this work.}
   ~~~~~~~ 
   Pedro Tabacof\samethanks
   ~~~~~~~ 
   Eduardo Valle \\
   RECOD Lab. — DCA / School of Electrical and Computer Engineering (FEEC)\\
   University of Campinas (Unicamp)\\
   Campinas, SP, Brazil \\
   \texttt{ \{roliveir, tabacof, dovalle\}@dca.fee.unicamp.br }
}

\begin{document}

\maketitle

\begin{abstract}
We evaluate the uncertainty quality in neural networks using anomaly detection. We extract uncertainty measures (e.g. entropy) from the predictions of candidate models, use those measures as features for an anomaly detector, and gauge how well the detector differentiates known from unknown classes. We assign higher uncertainty quality to candidate models that lead to better detectors. We also propose a3 novel method for sampling a variational approximation of a Bayesian neural network, called One-Sample Bayesian Approximation (OSBA). We experiment on two datasets, MNIST and CIFAR10. We compare the following candidate neural network models: Maximum Likelihood, Bayesian Dropout, OSBA, and --- for MNIST --- the standard variational approximation. We show that Bayesian Dropout and OSBA provide better uncertainty information than Maximum Likelihood, and are essentially equivalent to the standard variational approximation, but much faster.
\end{abstract}

\section{Introduction}

While current Deep Learning focuses on point estimates, many real-world applications require a full range of uncertainty. Reliable confidence on the prediction might be as useful as the prediction itself. The debate over the dangers of overconfident machine learning has reached the headlines of mass media~\cite{vaughan2016bpost, crawford2016ai}. Indeed, if our models are to drive cars, diagnose medical conditions, and even analyze the risk of criminal recidivism, unreliable confidence appraisal may have dire consequences. 

Traditional Deep Learning trains by maximum likelihood --- needing aggressive regularization to avoid overfitting --- and only provides point estimates, with limited uncertainty information. If the model outputs a vector of probabilities (as a softmax classifier does), we can quantify its uncertainty using the entropy of the prediction. However, the model can predict with high confidence for samples way outside the distribution seen during training~\cite{gal2016thesis}. Frequentist mitigations, like the bootstrap \cite{efron1994introduction}, do not scale well for deep models.

True Bayesian models infer the posterior distribution over all unknown factors, but their computational demands are often prohibitive. On the other hand, we may profitably reinterpret under a Bayesian perspective some of the ad hoc regularizations used in ordinary Deep Learning  (e.g., dropout~\cite{gal2015dropout, kingma2015variational}, early stopping~\cite{maclaurin2015early}, or weight decay~\cite{bishop2006pattern, blundell2015weight}). 
Gal and Ghahramani~\cite{gal2015dropout} show that multiple dropout forward passes in test time are equivalent to a Bayesian prediction (marginalized over the parameters' posteriors) given a particular variational approximation. A more direct (and  expensive) approach variationally approximates the posterior of each weight~\cite{blundell2015weight}. 



\section{One-Sample Bayesian Approximation (OSBA)}

Here we propose a novel Bayesian approach for neural networks, similar to the variational approximation of Blundell \emph{et al.}~\cite{blundell2015weight}, but much cheaper computationally. We call that approach \emph{One-Sample Bayesian Approximation} (OSBA), and investigate whether it achieves better quality of uncertainty information than traditional maximum likelihood.

We use exactly the same approach presented by Blundell \emph{et al.}~(\cite{blundell2015weight}, section 3.2), but instead of sampling the weight matrices for each training example, we sample the matrices only once per mini-batch, and use the same weights for all examples in that mini-batch. That approach leads to the same expected gradient, trading off higher variance for computational efficiency (about 10 times faster with a mini-batch of 100).

\section{Uncertainty Quality}

\begin{figure}[ht]
    \centering
    \includegraphics[width=12cm]{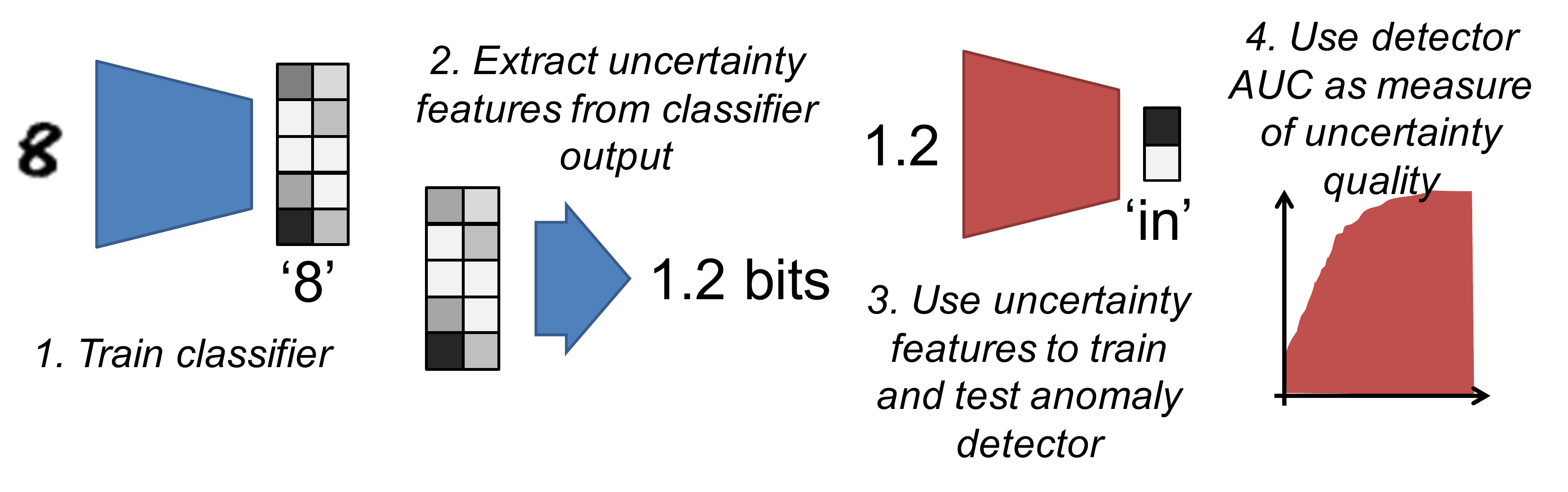}
    \caption{Uncertainty quality evaluation using an anomaly detection task. This is the experimental pipeline we follow to compare uncertainty quality among candidate models. (1) We train a candidate probabilistic classifier for the original task (MNIST or CIFAR10). (2) We extract uncertainty information from the classifier prediction. (3) We train a linear anomaly detector using those uncertainty measures as features. (4) We calculate the AUC of the anomaly detector. Higher detector AUCs indicate that a candidate model provides better uncertainty information.}
    \label{fig:pipeline}
\end{figure}

To evaluate the quality of uncertainty information, we employ anomaly detection: deciding whether or not a test sample belongs to the classes seen during training. More concretely, we pick a classification problem, exclude some classes from training, and use them to evaluate how much insight a candidate model has about its own classification confidence. We expect Bayesian neural networks to express such uncertainty well, to the point we can use it to decide whether a sample belongs or not to the known classes. Thus, we employ the AUC of the anomaly detector as a \emph{relative} measure of the quality of the uncertainty information output by candidate models (Figure~\ref{fig:pipeline}).

We contrast two experimental protocols. In the Blind Protocol, we separate the classes into two groups (In and Out); train the candidate neural network using only the In classes; and then train --- over the In \emph{vs.} Out classes --- a separate anomaly detector using the uncertainty extracted from the prediction of the candidate network. In the Calibrated Protocol, we separate the classes into three groups (In, Unknown, and Out); train the candidate network using the In classes with the loss function using the correct labels, and the Unknown classes with the loss function using the equiprobable prediction vector; and then train --- over the In \emph{vs.} Out classes --- a separate anomaly detector using the same features as before. The test set used to compute the AUC of the anomaly task excludes (obviously) all samples used to train the anomaly detector, and (perhaps less obviously) all samples used to train the candidate neural network.

\section{Methodology}

We use MNIST~\cite{lecun1998mnist} and CIFAR10~\cite{krizhevsky2009learning} datasets. For MNIST, the candidate networks have a two-layered fully-connected architecture with 512 neurons each, with dropout of 0.5 applied after each hidden layer. For CIFAR10, the candidate networks have two convolutional blocks (with dropout of 0.25 after each of them), followed by a fully-connected layer with 512 neurons (with dropout of 0.5). We optimize with ADAM~\cite{kingma2014adam}, and limit each training procedure to 100 epochs for MNIST, and 200 epochs for CIFAR10. For each dataset we choose 4 In classes, 4 Out classes, and (for the Calibrated Protocol) 2 Unknown classes (Table~\ref{table}). We randomize 20 combinations of In$\times$Out[$\times$Unknown] classes, with 5 repetitions each, totaling 100 replications.

\begin{table}[h]
\centering
\caption{Possible combination for In, Out and Unknown classes, showing one sample per class. MNIST's classes have crisp semantic separation; CIFAR10's have considerable overlap due to specialization (e.g., animals) or to background (e.g., sky, lawn, pavement). Such overlap might reduce the accuracy of anomaly detection as a measure of uncertainty quality.}
\label{table}
\begin{tabular}{M{2cm}M{4cm}M{4cm}M{2cm}}
\hline
Dataset & In & Out & Unknown \\ \hline
MNIST  &
\raisebox{-.7\height}{\includegraphics[width=0.8cm]{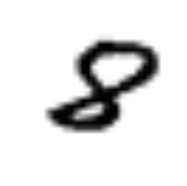}} \raisebox{-.7\height}{\includegraphics[width=0.8cm]{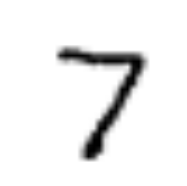}} \raisebox{-.7\height}{\includegraphics[width=0.8cm]{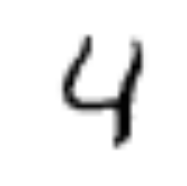}} \raisebox{-.7\height}{\includegraphics[width=0.8cm]{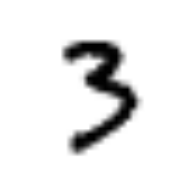}} & \raisebox{-.7\height}{\includegraphics[width=0.8cm]{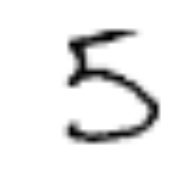}} \raisebox{-.7\height}{\includegraphics[width=0.8cm]{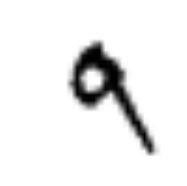}} \raisebox{-.7\height}{\includegraphics[width=0.8cm]{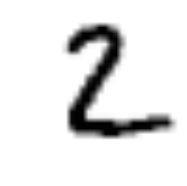}} \raisebox{-.7\height}{\includegraphics[width=0.8cm]{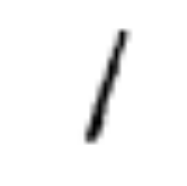}} & \raisebox{-.7\height}{\includegraphics[width=0.8cm]{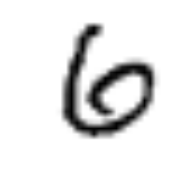}} \raisebox{-.7\height}{\includegraphics[width=0.8cm]{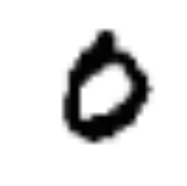}} 
\\ \hline
CIFAR  & 
\raisebox{-.7\height}{\includegraphics[width=0.8cm]{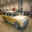}} \raisebox{-.7\height}{\includegraphics[width=0.8cm]{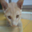}} \raisebox{-.7\height}{\includegraphics[width=0.8cm]{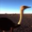}} \raisebox{-.7\height}{\includegraphics[width=0.8cm]{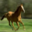}} & \raisebox{-.7\height}{\includegraphics[width=0.8cm]{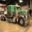}} \raisebox{-.7\height}{\includegraphics[width=0.8cm]{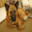}} \raisebox{-.7\height}{\includegraphics[width=0.8cm]{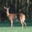}} \raisebox{-.7\height}{\includegraphics[width=0.8cm]{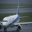}} & \raisebox{-.7\height}{\includegraphics[width=0.8cm]{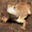}} \raisebox{-.7\height}{\includegraphics[width=0.8cm]{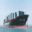}}
\\ \hline
\end{tabular}
\end{table}



As methods, we evaluate the usual baseline of Maximum Likelihood (ML), a Bayesian posterior estimated from dropout~\cite{gal2015dropout,gal2015bayesian} (BD), our approximation for the standard variational Bayesian neural networks using one sample per mini-batch (OSBA), and, for MNIST, we also evaluate the standard variational approximation~\cite{blundell2015weight} (SV).




The features for anomaly detection are uncertainty measures extracted from probabilistic predictions. For simplicity, the detector is a linear logistic classifier, with regularization parameter set by stratified cross-validation~\cite{refaeilzadeh2009cross}. For ML, only the vector of predicted probabilities is available, and thus we employ as feature the entropy --- the most theoretically sound measure of uncertainty --- over that vector. All Bayesian methods provide extra information; we use as feature vector the average and standard deviation of the entropy of the decision vector over 100 network prediction samples (estimating the expectation and variance of the entropy), the entropy of the average decision vector over those same samples (entropy of estimated expected predictions), and the average (over classes) of the standard deviations (over samples) of the predictions for each class.

\subsection{Bayesian ANOVA}

We analyze the results using Bayesian ANOVA~\cite{kruschke2014doing}, with a separate mean for each protocol (Blind \emph{vs.} Calibrated). That is equivalent to a two-way ANOVA without interactions, where the global mean and experimental protocol factors are fused together (for interpretability). The methods (ML, BD, OSBA) are the factors of variation. We constrain the sum of the effects to be zero, for identifiability. The response variable is the AUC of the anomaly detector. We use weakly informative priors. The following model reflects those choices:

\begin{align*}
model &= \{ML, BD, OSBA\} \\
protocol &= \{Blind, Calibrated\} \\
AUC_{protocol, model} &\sim \mathcal{N}(\mu_{protocol} + \theta_{model}, \sigma) \\
\mu_{protocol} &\sim \mathcal{N}(0, 10) \\
\theta_{model} &\sim \mathcal{N}(0, \sigma_{theta}) \\
\sigma_{theta} &\sim \text{Half-Cauchy}(0, 10) \\
\sigma &\sim \text{Half-Cauchy}(0, 10) \\
\sum_{i \in model} \theta_i &= 0
\end{align*}

We implement the model using Stan~\cite{carpenter2016stan}, and infer the posteriors of the unknown parameters using the NUTS algorithm~\cite{hoffman2014no}. To ensure proper convergence, we use 4 chains with 100K steps, including a 10K burn-in, and a thinning factor of 5. From Kruschke's suggestion~\cite{kruschke2014doing}, we present both the distribution of the marginal effects, and the distribution of the differences between effects.\footnote{Code for models, experiments, and analyses at \url{https://github.com/ramon-oliveira/deepstats}.}


\section{Results}

\begin{figure}[ht]
    \centering
    \includegraphics[width=3.2cm]{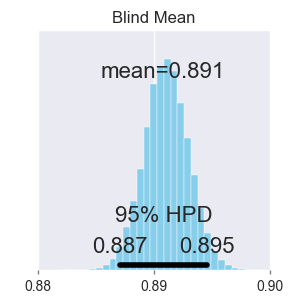}
    \includegraphics[width=3.2cm]{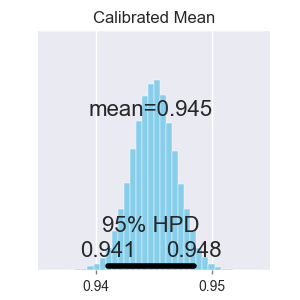} \\
    \includegraphics[width=3.2cm]{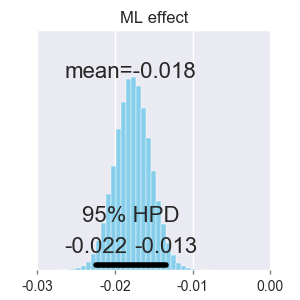} 
    \includegraphics[width=3.2cm]{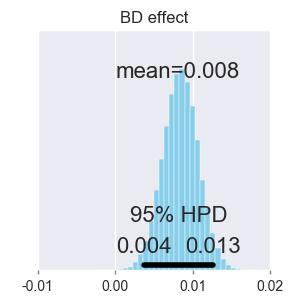}
    \includegraphics[width=3.2cm]{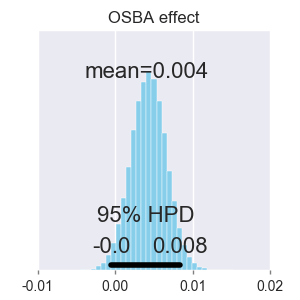} 
    \includegraphics[width=3.2cm]{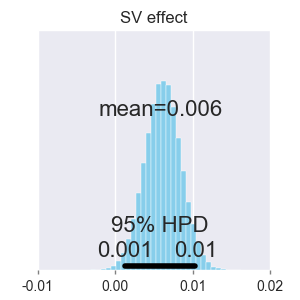} \\
    \includegraphics[width=3.2cm]{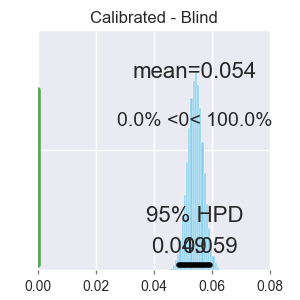}
	\includegraphics[width=3.2cm]{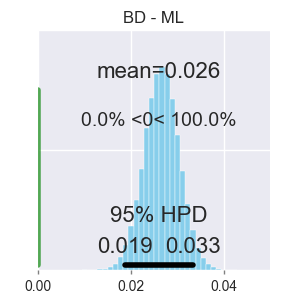}
    \includegraphics[width=3.2cm]{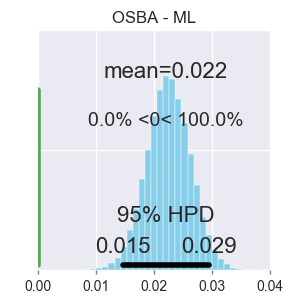}
    \includegraphics[width=3.2cm]{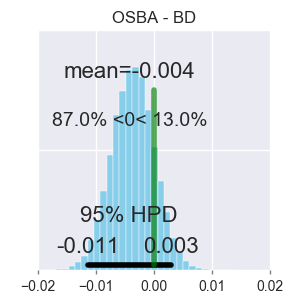}
    \includegraphics[width=3.2cm]{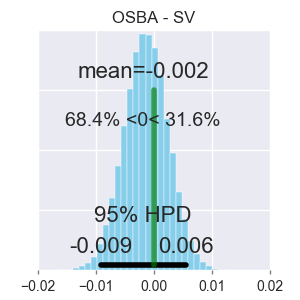}
    \includegraphics[width=3.2cm]{mnist/diff_os_drop.png}
    \includegraphics[width=3.2cm]{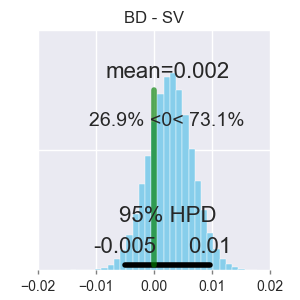}

    \caption{MNIST dataset. Each cell plots the distribution of the influence of the factor shown in the label above it, marginalized over all other factors. We highlight means (expected influence), and 95\% Highest Posterior Density intervals (HPD, black bars). On the topmost two rows, we consider the factors themselves (marginal effect), and on the other rows, the differences between effects. We consider the differences significant if the HPD does not contain 0.0 (green bar). The domain is the AUC of the anomaly detector.}
    \label{fig:mnist}
\end{figure}

\begin{figure}[ht]
    \centering
    \includegraphics[width=3.2cm]{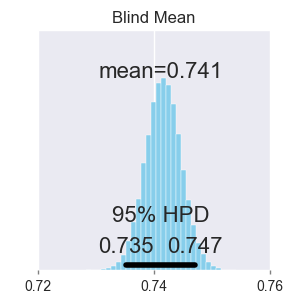}
    \includegraphics[width=3.2cm]{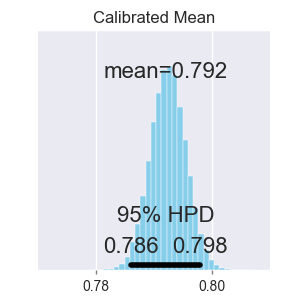} \\
    \includegraphics[width=3.2cm]{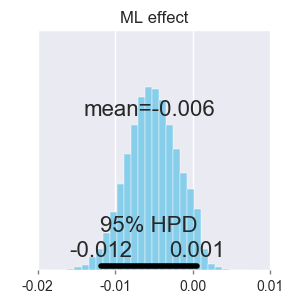} 
    \includegraphics[width=3.2cm]{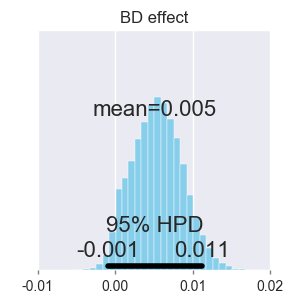}
    \includegraphics[width=3.2cm]{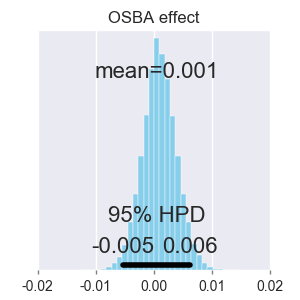}\\
    \includegraphics[width=3.2cm]{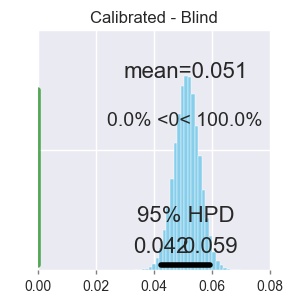}
	\includegraphics[width=3.2cm]{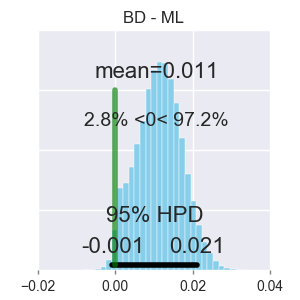}
    \includegraphics[width=3.2cm]{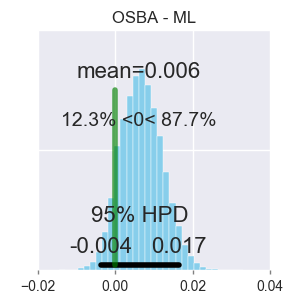}
    \includegraphics[width=3.2cm]{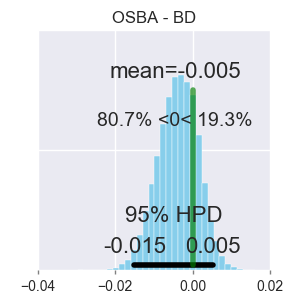}
    \caption{CIFAR10 dataset. Same information and interpretation as Figure~\ref{fig:mnist} above.}
    \label{fig:cifar10}
\end{figure}

\begin{figure}[ht]
    \centering
    \includegraphics[width=10cm]{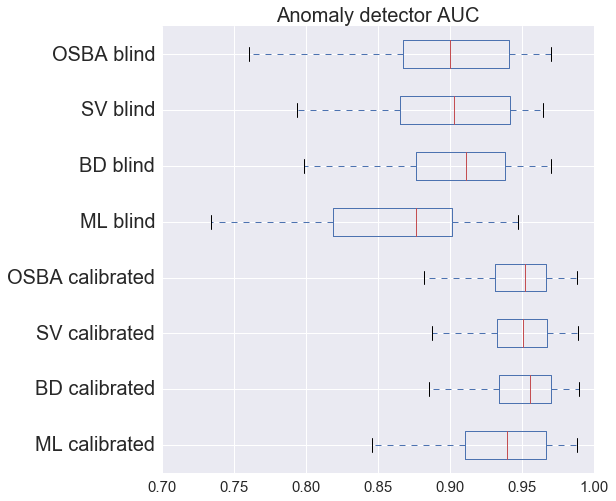}
    \caption{Distributions of the AUCs on MNIST for all combinations of probabilistic approach $\times$ experimental protocol. Each boxplot represents 100 replications, obtained by picking at random the In, Out, and  (for the calibrated protocol) Unknown classes.}
    \label{fig:mnist_anomaly}
\end{figure}

\begin{figure}[ht]
    \centering
    \includegraphics[width=10cm]{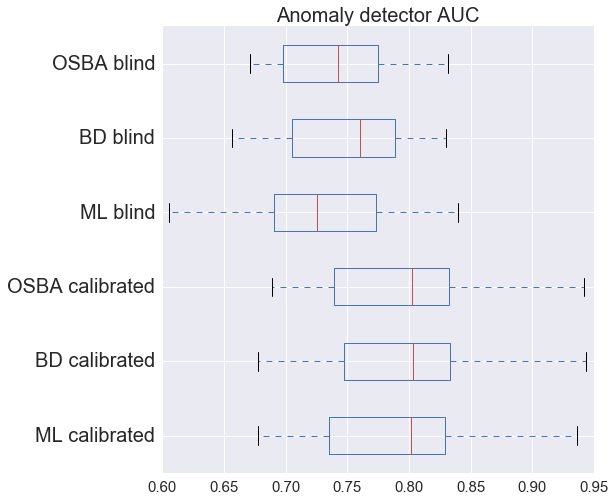}
    \caption{Distribution of the AUCs on CIFAR10, obtained the same way as Figure~\ref{fig:mnist_anomaly} above.}
    \label{fig:cifar10_anomaly}
\end{figure}

We show the Bayesian ANOVA results in Figures~\ref{fig:mnist} and~\ref{fig:cifar10} (for reference, we also show the raw distributions of the AUCs, as boxplots in Figures~\ref{fig:mnist_anomaly} and~\ref{fig:cifar10_anomaly}). Calibration with the auxiliary Unknown classes has a large effect, larger than choosing among uncertainty methods. Calibration, however, is not realistic for many applications, due to the artificial constraint of picking well-formatted Unknown classes. On the well-controlled scenario provided by MNIST, Bayesian methods give significantly better uncertainty information than ML. On MNIST, all Bayesian methods outperform ML, and their effects do not appear significantly different from each other. On CIFAR10, however, perfect semantic separation between classes is questionable (Table~\ref{table}), and the performance differences disappear: BD slightly outperforms ML, and OSBA slightly outperforms ML, but none of the differences appear significant.

\begin{table}[ht]
\centering
\caption{Test accuracy on the original classification task. We show the mean accuracy, with the standard deviation in parentheses, averaged over 100 different replications. Competing candidate models have similar accuracies, showing that enhanced uncertainty quality comes from enhanced probabilistic information, not from extra accuracy. Note that OSBA and SV have the same accuracy, but the latter is ten times slower.}
\label{testacc}
\begin{tabular}{l | l | l | l | l | l}
\hline
Dataset & Protocol   & ML & BD       & OSBA     & SV \\ \hline
MNIST   & Calibrated & 0.990 (0.002)      & 0.991 (0.002) & 0.991 (0.002) & 0.991 (0.002)    \\ MNIST   & Blind      & 0.992 (0.002)      & 0.992 (0.002) & 0.991 (0.002) & 0.991 (0.002)    \\ CIFAR10   & Calibrated & 0.878 (0.036)      & 0.896 (0.033) & 0.884 (0.037) & ---   \\ 
CIFAR10   & Blind      & 0.905 (0.029)      & 0.908 (0.028) & 0.896 (0.032) & ---    \\ 
\hline
\end{tabular}
\end{table}

Table~\ref{testacc} shows the accuracies of all candidate models. Note that competing candidate models have very similar performance: any gains in anomaly detection rather come from enhanced probabilistic information than from increased accuracy.

\section{Conclusion}

We formalized how to ascertain uncertainty quality of neural networks by using anomaly detection. We contrasted the usual maximum likelihood networks to Bayesian alternatives. Bayesian networks outperformed the frequentist network in all cases. 

We also proposed a novel way to sample from a variational approximation of a Bayesian neural network, OSBA, which is much faster than the standard sampling procedure, but still retains the same uncertainty quality. OSBA is 10$\times$ faster than SV; in our experiments, we observed relative training computational costs of 1$\times$ (ML) to 1$\times$ (BD) to 3$\times$ (OSBA) to 30$\times$ (SV). 

We believe, thus, that techniques like BD and OSBA deserve further investigation in more contexts. Finding a general measure of uncertainty quality is, however, still a challenge. Our experiments suggest that anomaly detection only gives good uncertainty measures for well-separated classes, like MNIST's; for uncontrolled datasets like CIFAR10 (or ImageNet), we need a measure that tolerates a degree of semantic intersection between the classes.

As future work, we intend to explore other forms of uncertainty quality evaluation, and to test OSBA in more varied settings.

\subsubsection*{Acknowledgments}

We thank Brazilian agencies CAPES, CNPq and FAPESP for financial support. We gratefully acknowledge the support of NVIDIA Corporation with the donation of the Tesla K40 GPU used for this research. Eduardo Valle is partially supported by a Google Awards LatAm 2016 grant, and by a CNPq PQ-2 grant (311486/2014-2). Ramon Oliveira is supported by a grant from Motorola Mobility Brazil.

\bibliographystyle{unsrt}
\bibliography{references.bib}

\begin{thebibliography}{10}

\bibitem{vaughan2016bpost}
Jennifer Vaughan and Hanna Wallach.
\newblock The inescapability of uncertainty: Ai, uncertainty, and why you
  should vote no matter what predictions say.
\newblock \url{https://medium.com/@jennwv/uncertainty-edd5caf8981b}, 2016.

\bibitem{crawford2016ai}
Kate Crawford.
\newblock Artificial intelligence's white guy problem.
\newblock {\em The New York Times}, 2016.

\bibitem{gal2016thesis}
Yarin Gal.
\newblock {\em Uncertainty in Deep Learning}.
\newblock PhD thesis, University of Cambridge, 2016.

\bibitem{efron1994introduction}
Bradley Efron and Robert~J Tibshirani.
\newblock {\em An introduction to the bootstrap}.
\newblock CRC press, 1994.

\bibitem{gal2015dropout}
Yarin Gal and Zoubin Ghahramani.
\newblock Dropout as a bayesian approximation: Representing model uncertainty
  in deep learning.
\newblock {\em arXiv preprint arXiv:1506.02142}, 2015.

\bibitem{kingma2015variational}
Diederik~P Kingma, Tim Salimans, and Max Welling.
\newblock Variational dropout and the local reparameterization trick.
\newblock {\em arXiv preprint arXiv:1506.02557}, 2015.

\bibitem{maclaurin2015early}
Dougal Maclaurin, David Duvenaud, and Ryan~P Adams.
\newblock Early stopping is nonparametric variational inference.
\newblock {\em arXiv preprint arXiv:1504.01344}, 2015.

\bibitem{bishop2006pattern}
Christopher~M Bishop.
\newblock Pattern recognition.
\newblock {\em Machine Learning}, 128, 2006.

\bibitem{blundell2015weight}
Charles Blundell, Julien Cornebise, Koray Kavukcuoglu, and Daan Wierstra.
\newblock Weight uncertainty in neural network.
\newblock In {\em Proceedings of The 32nd International Conference on Machine
  Learning}, pages 1613--1622, 2015.

\bibitem{lecun1998mnist}
Yann LeCun, Corinna Cortes, and Christopher~JC Burges.
\newblock The mnist database of handwritten digits, 1998.

\bibitem{krizhevsky2009learning}
Alex Krizhevsky and Geoffrey Hinton.
\newblock Learning multiple layers of features from tiny images.
\newblock 2009.

\bibitem{kingma2014adam}
Diederik Kingma and Jimmy Ba.
\newblock Adam: A method for stochastic optimization.
\newblock {\em arXiv preprint arXiv:1412.6980}, 2014.

\bibitem{gal2015bayesian}
Yarin Gal and Zoubin Ghahramani.
\newblock Bayesian convolutional neural networks with bernoulli approximate
  variational inference.
\newblock {\em arXiv preprint arXiv:1506.02158}, 2015.

\bibitem{refaeilzadeh2009cross}
Payam Refaeilzadeh, Lei Tang, and Huan Liu.
\newblock Cross-validation.
\newblock In {\em Encyclopedia of database systems}, pages 532--538. Springer,
  2009.

\bibitem{kruschke2014doing}
John Kruschke.
\newblock {\em Doing Bayesian data analysis: A tutorial with R, JAGS, and
  Stan}.
\newblock Academic Press, 2014.

\bibitem{carpenter2016stan}
Bob Carpenter, Andrew Gelman, Matt Hoffman, Daniel Lee, Ben Goodrich, Michael
  Betancourt, Michael~A Brubaker, Jiqiang Guo, Peter Li, and Allen Riddell.
\newblock Stan: A probabilistic programming language.
\newblock {\em J Stat Softw}, 2016.

\bibitem{hoffman2014no}
Matthew~D Hoffman and Andrew Gelman.
\newblock The no-u-turn sampler: adaptively setting path lengths in hamiltonian
  monte carlo.
\newblock {\em Journal of Machine Learning Research}, 15(1):1593--1623, 2014.

\end{thebibliography}


\end{document}